\documentclass[a4paper,10pt,onecolumn,notitlepage,final]{article}
\usepackage{graphicx}
\usepackage{bm}
\usepackage{booktabs}

\begin{document}
\title{Swarm Robots Inspired by Friendship Formation Process}
\maketitle
\author{Takeshi Kano$^{\dagger}$$^{\ast}$, Naoki matsui$^{\dagger}$, Eiichi Naito$^{\dagger\dagger}$, Takenobu Aoshima$^{\dagger\dagger}$, and Akio Ishiguro$^{\dagger}$\vspace{5mm}\\ 
$^{\dagger}${\em{Research Institute of Electrical Communication, Tohoku University, 2-1-1 Katahira, Aobaku, Sendai 980-8577, Japan}}\\
$^{\dagger\dagger}${\em{Business Innovation Division, Panasonic Corporation, 1006
Kadoma, Kadoma City, Osaka 571-8501, Japan}}\\
\thanks{$^\ast$Contact author. Tel: +81-22-217-5465. Email: tkano@riec.tohoku.ac.jp}

\begin{abstract}
Swarm robotic systems are systems in which multiple robots having simple functionality perform tasks through their cooperation, and are advantageous in that they can exhibit non-trivial macroscopic functions such as adaptability, fault tolerance, and scalability. We previously proposed a simple model of swarm formation inspired by friendship formation process in human society, and demonstrated via simulation that various non-trivial patterns emerge. In this study, we examine the applicability of the proposed model to a swarm robotic system. As a first step, we developed five robots and demonstrated via real-world experiments that the simulation results can be largely reproduced.
\end{abstract}


\section{Introduction}
Swarm formation through local interaction among individuals is widely found in natural and social systems such such as the flocking of animals \cite{Calovi,Chen,Cristiani,Couzin1,Couzin2,Gregoire,Hayakawa,Levine,Motsch,Olson,Pearce,Reynolds,Romanczuk,Vicsek,Weitz,Zheng}, traffic and pedestrian flow \cite{Bando,Helbing}, and social networks \cite{Gonzalez,Kano-NOLTA}. An interesting aspect of swarms is that nontrivial macroscopic functions such as adaptability, scalability, and fault tolerance emerge although each individual has only trivial functions. Thus, swarms have attracted attention to engineers, and many swarm robotic systems have been developed thus far \cite{Navarro,Tan,Deneubourg,Kurabayashi}.

Recently, we have proposed an extremely simple mathematical model of swarm formation \cite{Kano-ECAL,Kano-JPSJ}. This model was inspired by the friendship formation process in human society (for example,  a process in which several cliques are formed spontaneously in certain communities such as classes in schools). It was demonstrated via simulations of the proposed model that various patterns emerge by changing the parameters (Fig. 1). Some of the patterns are dynamic and lifelike, and it was found that non-reciprocal property of the interaction between agents plays a crucial role for the emergence of these patterns. This model has many possible applications that range from science to engineering, such as understanding the core principle of self-organization \cite{Kano-ECAL}, elucidating the essential mechanism of the behavior of active matters \cite{Tanaka-JPSJ}, and designing swarm robotic systems \cite{Kano-Swarm}.

In this study, we focus on the application of the proposed model to the design of swarm robotic systems. Specifically, hardware implementation of the proposed model is discussed. We developed swarm robots, each of which can move omni-directionally and can detect relative position between itself and nearby robots. We performed real-world experiments by using five robots, and demonstrated that the simulation results can be largely reproduced.

This paper is organized as follows. In section 2, our model proposed previously is reviewed briefly. In section 3, we introduce the hardware design of the robot. In section 4, experimental results by using the developed robots are presented. Limitation of the developed robot is also discussed. In section 5, the conclusions and recommendations for future studies are presented.

\vspace{-4mm}
\section{Review of our previous works}
\vspace{-2mm}
Let us briefly summarize the model which we proposed previously \cite{Kano-ECAL,Kano-JPSJ}. Particles, each of which represents a person in a community, exist on a two-dimensional plane, and the position of the $i$th particle ($i=1,2,\cdot\cdot\cdot, N$) is denoted by $\mathbf{r}_i$. The time evolution of $\mathbf{r}_i$ is given by
\begin{equation}
\dot{\mathbf{r}}_i=\sum_{j\neq i}(k_{ij}|\mathbf{R}_{ij}|^{-1}-|\mathbf{R}_{ij}|^{-2})\hat{\mathbf{R}}_{ij}, \label{1.1}
\vspace{-3.0mm}
\end{equation}
where $\mathbf{R}_{ij}=\mathbf{r}_j-\mathbf{r}_i$, $\hat{\mathbf{R}}_{ij}=\mathbf{R}_{ij}/|\mathbf{R}_{ij}|$, and $k_{ij}$ denotes a constant that represents ``to what extent person $i$ prefers person $j$." When $k_{ij}=k_{ji}$, the interaction between the $i$th and $j$th particles is described by a potential, and the distance between the $i$th and $j$th particles tends to converge to $k_{ij}^{-1}$ (if $k_{ij}>0$). However, because $k_{ij}$ is not necessarily equal to $k_{ji}$, {\em i.e.}, the interaction can be non-reciprocal, Eq. (\ref{1.1}) is generally a non-equilibrium open system in which both energy and momentum are non-conservative.

Simulation results for several $k_{ij}$ values can be downloaded from\\
http://www.riec.tohoku.ac.jp/$\sim$tkano/ECAL$\_$Movie1.mp4. \\
Several examples are shown in Fig. \ref{fig1}. It is found that various nontrivial patterns emerge. 

\begin{figure}[t]
\centering
\includegraphics[width=8cm]{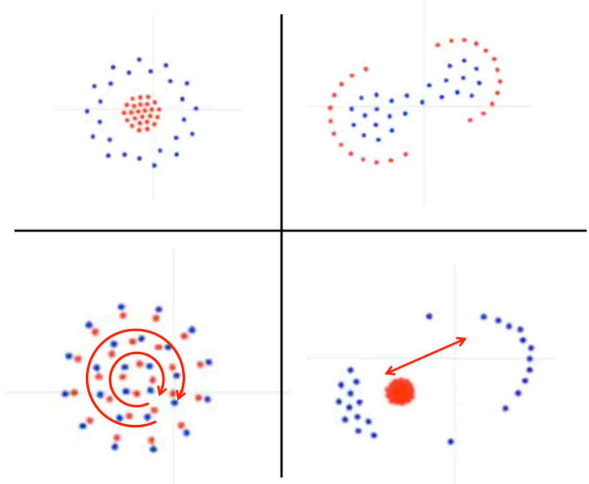}
\vspace{-2mm}
\caption{Examples of simulation results of the model proposed in ref. \cite{Kano-ECAL,Kano-JPSJ}.}
\label{fig1}
\vspace{-2mm}
\end{figure}

\section{Robot}
\vspace{-2mm}
It is not easy to implement the proposed model in a hardware, because detecting the $k_{ij}$ values for arbitral parameter sets is technically difficult. Hence, in this study, we developed robots applicable only to the case where $N=5$ and $k_{ij}=k_p+k_m~(i=1, 2\leq j\leq 5)$, $k_p-k_m~(2\leq i\leq 5, j=1)$ and $k_a~(2\leq i\leq 5, 2\leq j\leq 5)$,
where $k_p$, $k_m$, and $k_a$ are constants. This case was studied previously by simulations and mathematical analyses, and it is already known that versatile patterns, {\em e.g.}, static configuration, translational motion with relative position among particles fixed, and limit cycle oscillation, emerge by changing the parameter values \cite{Kano-JPSJ}.

\begin{figure}[t]
\centering
\includegraphics[width=7cm]{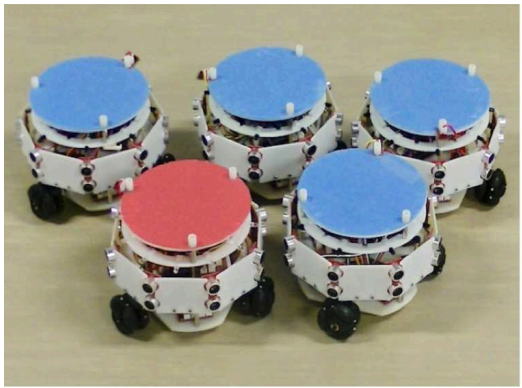}
\caption{Overview of the robots. }
\label{fig2}
\vspace{-2mm}
\end{figure}

\begin{figure}[t]
\centering
\includegraphics[width=8.5cm]{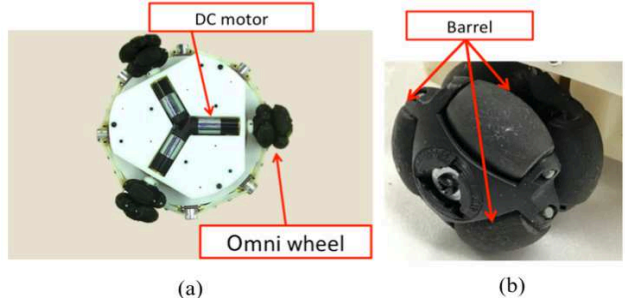}
\caption{Detailed structure of the robot: (a) bottom view and (b) magnified view of an omni-wheel. }
\label{fig3}
\vspace{-2mm}
\end{figure}

\begin{figure}[t]
\centering
\includegraphics[width=5cm]{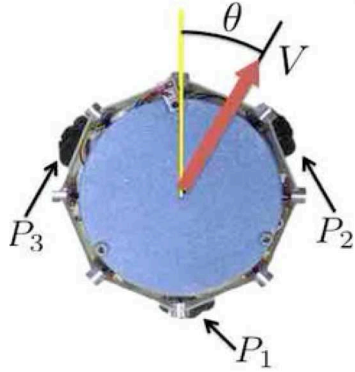}
\caption{Definition of $V$, $\theta$, $P_1$, $P_2$, and $P_3$.}
\label{fig4}
\vspace{-2mm}
\end{figure}

\begin{figure}[t]
\centering
\includegraphics[width=6.5cm]{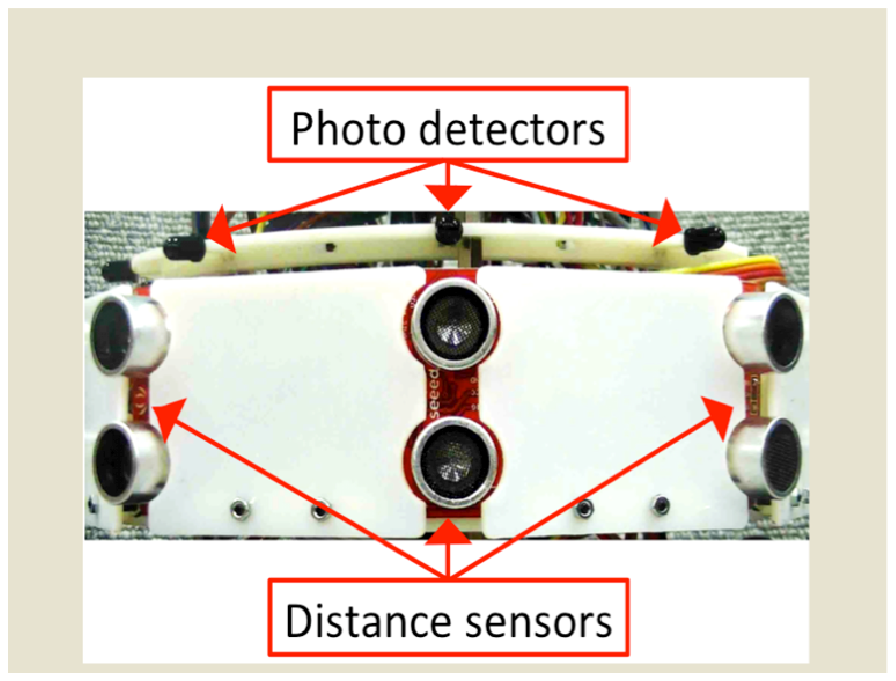}
\caption{Distance sensors and photodetectors. }
\label{fig5}
\vspace{-3mm}
\end{figure}

Figure 2 shows the overview of the robots. Each robot is cylindrical shape, and its diameter and mass are 0.19 m and 1.2 kg, respectively. Three omni-wheels (TYPE 2571, Tosa densi Co.,Ltd., slightly modified) are attached equidistantly at the bottom (Fig. 3(a)). Each omni-wheel consists of a couple of discs (Fig. 3(b)), and they rotate owing to the torque generated by a DC motor (Maxon Japan Corporation, RE-max17 GB 4.5W SL 2WE) (Fig. 3(a)). The motor axes are in parallel to the radial direction of the robot. Three barrels are implemented equidistantly in each disc so that the wheel can passively move in the direction parallel to the motor axis (Fig. 3(b)). 

Thus, the robot can move omni-directionally by changing the output ratio of the motors. Specifically, the motor outputs $P_1$, $P_2$, and $P_3$ were calculated from the direction of motion $\theta$ and the velocity of the robot $V$ as follows (Fig. 4):
\begin{eqnarray}
\left[\begin{array}{@{\,}ccc@{\,}}
P_1 \\
P_2 \\
P_3 \\
\end{array}
\right] =c \left[\begin{array}{@{\,}ccc@{\,}}
0 & -1  \\
-\sqrt{3}/2 & 1/2\\
\sqrt{3}/2 & 1/2\\
\end{array}
\right] \left[\begin{array}{@{\,}cc@{\,}}
V\cos\theta \\
 V\sin\theta\\
\end{array}
\right], 
\end{eqnarray} 
where $c$ is a positive constant. 

Eight pairs of ultrasonic distance sensor modules (101990004, Speed Studio Co.) and photodetector (SID1K10CM, Linkman Co.) are attached equidistantly on the side surface of the robot (Fig. 5). Each distance sensor can detect robots within 1.7 m and $\pm \pi/12$ rad from the direction it points. The sensor value was updated every 0.1 second. Thus, the outputs of the distance sensors enable the robot to identify the relative position of its neighboring robots with respect to itself in most cases. Infrared light LED is attached to one of the robots ($i=1$). The robots can identify the $k_{ij}$ values of the neighboring robots by detecting the infrared light via the photodetectors ({\em i.e.}, $j=1$ if the $i$th robot detects the infrared from the $j$th robot, otherwise $2\leq j\leq 5$). 

A microcomputer (mbed:NXP LPC 1768) is embedded in each robot to determine its direction of motion and velocity on the basis of the sensory information obtained. The values of $k_p$, $k_m$, and $k_a$ can be changed via wireless communication (Programmable XBee ZB(S2C)/Wire antenae type, Digi Co.)

\vspace{-2mm}
\section{Experimental results}
\vspace{-2mm}
\begin{figure}[t]
\centering
\includegraphics[width=8.5cm]{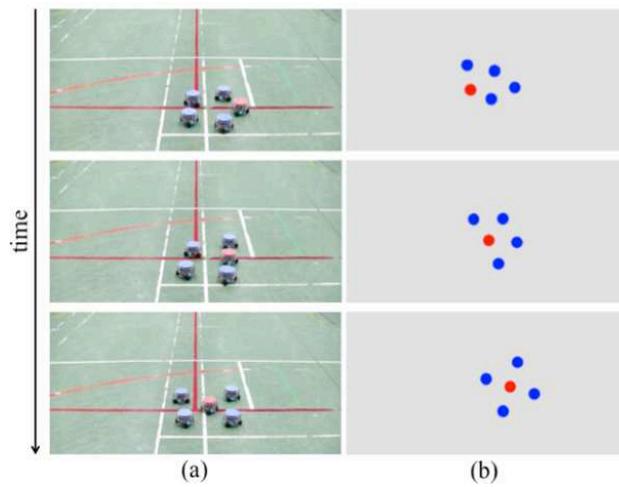}
\caption{Result for $k_p=2.0$, $k_m=0.5$, and $k_a=2.0$: (a) Robot experiment and (b) Simulation.}
\label{fig6}
\end{figure}

\begin{figure}[t]
\centering
\includegraphics[width=8.5cm]{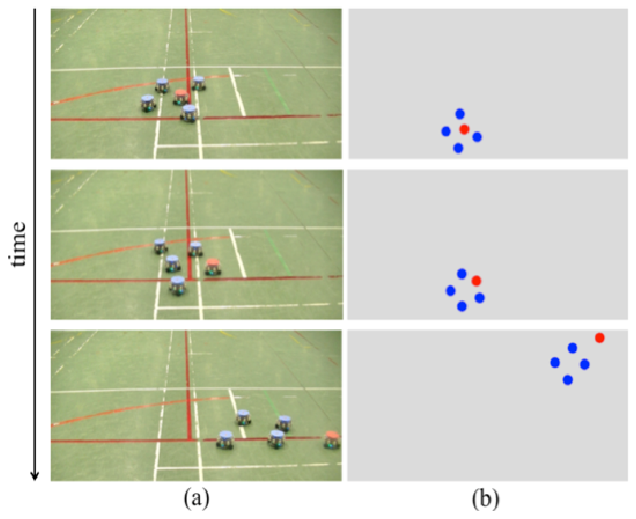}
\caption{Result for $k_p=2.0$, $k_m=-1.0$, and $k_a=2.0$: (a) Robot experiment and (b) Simulation.}
\label{fig7}
\end{figure}

\begin{figure}[t]
\centering
\includegraphics[width=8.5cm]{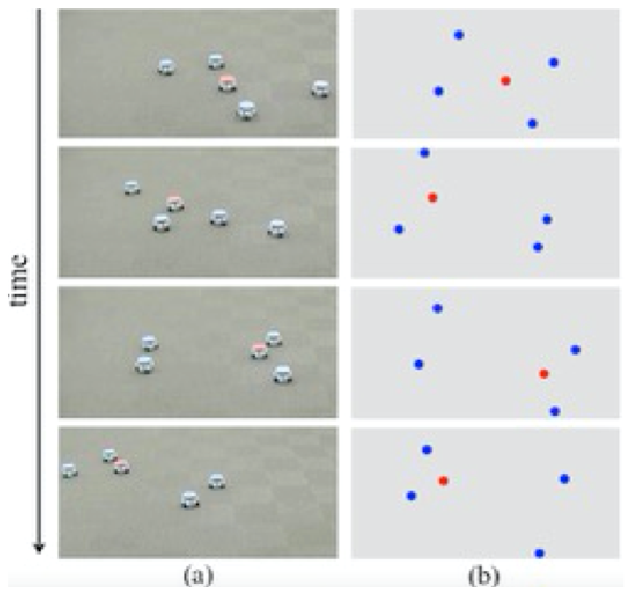}
\caption{Result for $k_p=1.6$, $k_m=2.4$, and $k_a=1.6$: (a) Robot experiment and (b) Simulation.}
\label{fig8}
\end{figure}

We implemented the proposed model (Eq. (1)) in the robots developed. The color of the robot was red for $i=1$ and blue for $2\leq i\leq 5$ (Fig. 2). Figure 6(a) shows the behavior of the robots when $k_p = 2.0$, $k_m=0.5$, $k_a=2.0$. Although the robots formed pentagonal shape at the initial condition, the configuration converged to a stationary state in which the red robot was surrounded by the blue robots. This is in good agreement with the simulation result (Fig. 6(b)). Figure 7 shows the results when $k_p = 2.0$, $k_m=-1.0$, $k_a=2.0$. The red robot was surrounded by the blue robots at the initial condition. The robots exhibited translational motion whereby the red robot was chased by the blue robots (Fig. 7(a)). This behavior qualitatively agrees with the simulation result (Fig. 7(b)). 

Finally, we performed experiments with $k_p=1.6$, $k_m=2.4$, $k_a=1.6$. The red robot was surrounded by the blue robots at the initial condition. The red robot moved periodically with surrounded by the blue robots (Fig. 8(a)). This behavior is similar to the simulation result (Fig. 8(b)). However, the behavior of the robots is not completely periodic but somewhat irregular, compared with the simulation result. 

The discrepancy between the robot experiment and the simulation are considered to be due to the limitation of the function of the sensors implemented in the robots. For example, the sensor range is limited, and the time interval for the update of the sensor value is not short enough. Further, noise in the sensor output cannot be neglected. These problems still need to be solved.

\section{Conclusion and future works}
We developed swarm robots based on the minimal model inspired by friendship formation process \cite{Kano-ECAL,Kano-JPSJ}. As a first step, we investigated whether five developed robots behave in similar manners as the simulation results. As a result, we largely succeeded in reproducing the simulation results. Our results indicate that the proposed model can be applied to swarm robotic systems.

There still exists a discrepancy between the robot experiment and the simulation. Further, the robots developed in this study can be applicable only to specific $k_{ij}$ values. Solving these problems through the improvement of the hardware remains as a future work.

\end{document}